\documentclass{article}

\PassOptionsToPackage{numbers, compress}{natbib}

\usepackage[final]{neurips}

\usepackage{latexsym}
\usepackage{tcolorbox}

\usepackage[T1]{fontenc}
\usepackage[utf8]{inputenc}
\usepackage{microtype}

\usepackage{inconsolata}
\usepackage{wrapfig,lipsum,booktabs}
\usepackage{pifont}
\usepackage{makecell} 
\usepackage{tabularx}
\usepackage[normalem]{ulem}
\useunder{\uline}{\ul}{}
\usepackage[export]{adjustbox}
\usepackage[table]{xcolor}

\usepackage{hyperref}       
\usepackage{url}            
\usepackage{booktabs}       
\usepackage{amsfonts}       
\usepackage{nicefrac}       

\usepackage{graphicx}
\usepackage{subfigure}
\usepackage{float}
\usepackage{multirow}

\usepackage{listings}

\usepackage{amsmath}
\usepackage{amssymb}
\usepackage{mathtools}
\usepackage{amsthm}
\usepackage{balance}
\usepackage{flushend}

\usepackage{lipsum}
\usepackage{graphicx}
\usepackage{subcaption}
\usepackage{tikz}
\usetikzlibrary{shapes, arrows, positioning}

\usepackage{fancyhdr}

\hypersetup{
	colorlinks,
	citecolor=gray,
	linkcolor=red,
	urlcolor=blue}

\usepackage[capitalize,noabbrev]{cleveref}
\usepackage{tablefootnote}

\usepackage{color, colortbl,xcolor}
\usepackage{enumitem}
\usepackage{booktabs,arydshln}

\definecolor{iadablue}{RGB}{235,243,255}
\usepackage{xspace}
\usepackage{natbib}
\usepackage{doi}

\newcommand{\nbf}[1]{{\noindent \textbf{#1}}}

\def\paperTitle{Mitigating the Reasoning Tax in Vision-Language Fine-Tuning with Input-Adaptive Depth Aggregation}

\title{\paperTitle}

\makeatletter
\def\adl@drawiv#1#2#3{%
	\hskip.5\tabcolsep
	\xleaders#3{#2.5\@tempdimb #1{1}#2.5\@tempdimb}%
	#2\z@ plus1fil minus1fil\relax
	\hskip.5\tabcolsep}
\newcommand{\cdashlinelr}[1]{%
	\noalign{\vskip\aboverulesep
		\global\let\@dashdrawstore\adl@draw
		\global\let\adl@draw\adl@drawiv}
	\cdashline{#1}
	\noalign{\global\let\adl@draw\@dashdrawstore
		\vskip\belowrulesep}}
\makeatother

\setlist[itemize]{align=parleft,left=0pt..0.5em}
\setlist[enumerate]{align=parleft,left=0pt..0.5em}

\setlist[itemize]{align=parleft,left=0pt..0.8em}

\usepackage{booktabs}
\newcolumntype{g}{>{\columncolor{airforceblue}}c}
\def\authorBlock{
    Yiming Ren \qquad
    Yujiu Yang\footnotemark[1] \qquad
    Junjie Wang\footnotemark[1]
    \\\\\fontsize{10}{10}
    \selectfont{Tsinghua University} \\
    {\tt\small rym24@mails.tsinghua.edu.cn}, \quad
    {\tt\small \{yang.yujiu, wangjunjie\}@sz.tsinghua.edu.cn}
}

\author{\authorBlock}

\newcommand{\method}{IADA\xspace}
\newcommand{\attnres}{AttnRes\xspace}

\definecolor{lightgray}{gray}{0.92}

\begin{document}

\maketitle

{
  \renewcommand{\thefootnote}%
  {\fnsymbol{footnote}}
  \footnotetext[1]{Corresponding Author.}
}

\begin{abstract}

Supervised fine-tuning (SFT) on visual instruction data often improves perceptual capabilities in vision-language models (VLMs) while degrading reasoning performance, creating a persistent reasoning tax during post-training. 
We investigate whether this degradation is related to disrupted access to depth-wise representations, and find that even fixed cross-depth aggregation substantially restores reasoning, suggesting that preserved cross-depth access is an important missing factor in VLM fine-tuning. 
Building on this observation, we propose Input-Adaptive Depth Aggregation (IADA), a lightweight mechanism that makes cross-depth retrieval input-adaptive, modality-aware, and efficiently parameterized through a low-rank bottleneck. 
On Qwen3-VL-2B, IADA improves the average reasoning score by $9.5$ points and the average perception score by $3.3$ points over LoRA-only fine-tuning with only $0.14$M additional parameters, with the strongest gains appearing in parameter-efficient low-rank settings.


\end{abstract}

\section{Introduction}
\label{sec:intro}


In recent years, vision-language models (VLMs) have achieved substantial progress on tasks such as OCR, chart understanding, and multimodal instruction following~\cite{liu2024visual, chen2024internvl, yang2024qwen2, bai2025qwen2vl}. 
However, an increasingly prominent phenomenon is that supervised fine-tuning on visual instruction data, while improving perceptual capabilities, often weakens reasoning performance, particularly on mathematical, scientific, and complex visual inference tasks~\cite{luo2024empirical, kotha2024understanding}. 
On Qwen3-VL-2B, we observe that LoRA fine-tuning reduces the average reasoning score by $5.1$ points while yielding only limited gains in perception (details in~\cref{tab:main_results}). 
This phenomenon can be viewed as a form of ``reasoning tax'' in VLM post-training, yet its underlying mechanism remains poorly understood.


Motivated by this perspective, we hypothesize that \textit{VLMs acquire a depth-wise feature hierarchy during pretraining}. 
Earlier layers tend to encode low-level perceptual patterns, whereas deeper layers encode more abstract reasoning representations~\cite{veit2016residual}. 
Uniform supervised fine-tuning may disturb this hierarchy. 
If this is the case, the model may no longer access reasoning-relevant representations as effectively.

To test this idea, we first introduce fixed-query attention residuals~\cite{team2025kimi} as a minimal intervention. 
Simply restoring cross-depth access raises the average reasoning score from $52.2$ to $58.7$. 
Although this result is not a definitive mechanistic proof, it provides strong empirical support for the role of cross-depth information access in the VLM reasoning tax.
Fixed aggregation, however, is still limited. 
It uses the same retrieval strategy for every input and treats all modalities identically. 
We therefore study the design space of depth aggregation more systematically. 
The results show that input-adaptive query generation is important for further reasoning recovery. 
They also show that useful routing behavior can be captured with a highly compact low-rank bottleneck. 
A rank-16 bottleneck with $0.14$M parameters matches a full-rank projection with $8.4$M parameters. 
We further find that self-modal conditioning is more effective than cross-modal conditioning for depth-routing decisions.

Building on these observations, we propose Input-Adaptive Depth Aggregation (IADA). 
IADA inserts lightweight modules at transformer block boundaries. 
These modules generate modality-specific aggregation queries from the current input and perform efficient cross-depth retrieval through low-rank projections. 
Unlike conventional methods such as LoRA, which primarily modify transformations within individual layers, IADA explicitly models across-layer routing during fine-tuning. 
It therefore serves as a structural complement to conventional PEFT.
On Qwen3-VL-2B, combining IADA with LoRA yields the strongest overall performance among all baselines, raising the average reasoning score to $61.7$ and the average perception score to $75.3$ with only $0.14$M additional parameters.
Further ablations show that these gains are most pronounced under low-rank LoRA settings. 
This finding suggests that explicit depth routing is particularly effective in capacity-constrained parameter-efficient fine-tuning regimes.

\nbf{Our contributions are threefold.}
\begin{enumerate}[leftmargin=1.5em,itemsep=0pt,topsep=2pt]
\item We reinterpret the reasoning tax in VLM fine-tuning as a problem of cross-depth representational access, and provide empirical support for this view through the recovery achieved by fixed depth aggregation.
\item We propose \method, which explicitly models cross-depth information routing in an input-adaptive, modality-aware, and low-rank manner.
\item Through systematic experiments, we show that the gains of \method primarily come from input-adaptive depth routing, and that a very small low-rank bottleneck is sufficient to capture most of the useful aggregation behavior.
\end{enumerate}

Overall, our results highlight cross-depth information routing as an important and independent design axis in multimodal fine-tuning beyond conventional within-layer adaptation.

\section{Related Work}
\label{sec:related}

\nbf{Reasoning Degradation under SFT and Instruction Tuning.}
A growing body of work shows that supervised fine-tuning can improve task-specific behavior while degrading reasoning performance in both LLMs~\cite{ouyang2022training, kirkpatrick2017overcoming} and VLMs~\cite{luo2024empirical, kotha2024understanding}. 
Existing mitigations mainly focus on the fine-tuning process itself, including data mixing, regularization, and distillation. 
These approaches treat reasoning degradation primarily as a training-time stability problem. In contrast, we study it from a structural perspective. 
Our hypothesis is that fine-tuning may disrupt access to representations distributed across depth, and that restoring such access can mitigate the reasoning tax.

\nbf{Parameter-Efficient Adaptation in LLMs and VLMs.}
Parameter-efficient fine-tuning methods, including LoRA~\cite{hu2022lora}, QLoRA~\cite{dettmers2024qlora}, adapters~\cite{houlsby2019parameter, chen2022adaptformer}, and prompt tuning~\cite{lester2021power, li2021prefix}, primarily operate within individual layers. 
Their central question is how to inject or optimize a small number of parameters while preserving the pretrained backbone. 
This layer-local view also dominates adaptation design in VLMs. Our method addresses an orthogonal question: how fine-tuning should preserve or reconfigure information flow across layers. 
IADA therefore complements conventional PEFT rather than replacing it.

\nbf{Cross-Depth Aggregation and Residual Access.}
Cross-depth feature reuse has a long history in deep architectures. DenseNet~\cite{huang2017densely} and Deep Layer Aggregation~\cite{yu2018deep} showed the value of aggregating representations across depth, and later VLM systems such as DeepStack~\cite{chen2024deepstack} also reused visual features from multiple layers. 
The most relevant recent work is Attention Residuals (\attnres) in Kimi k1.5~\cite{team2025kimi}, which introduces attention-based residual access across layer blocks using fixed queries. 
Our work builds on this line but focuses on a different setting and a different question: VLM fine-tuning under reasoning degradation. 
In this setting, we show that effective depth aggregation should be input-adaptive and modality-aware. 
More broadly, this design choice is compatible with modern VLMs, which already rely on shared self-attention for cross-modal interaction~\cite{liu2024visual, chen2024internvl, yang2024qwen2, bai2025qwen2vl}. 
Our results suggest that while shared attention can handle token-level information exchange, depth-routing decisions benefit from modality-local conditioning.

\section{Method}
\label{sec:method}

%






\begin{figure}[t]
\centering
\includegraphics[width=\linewidth]{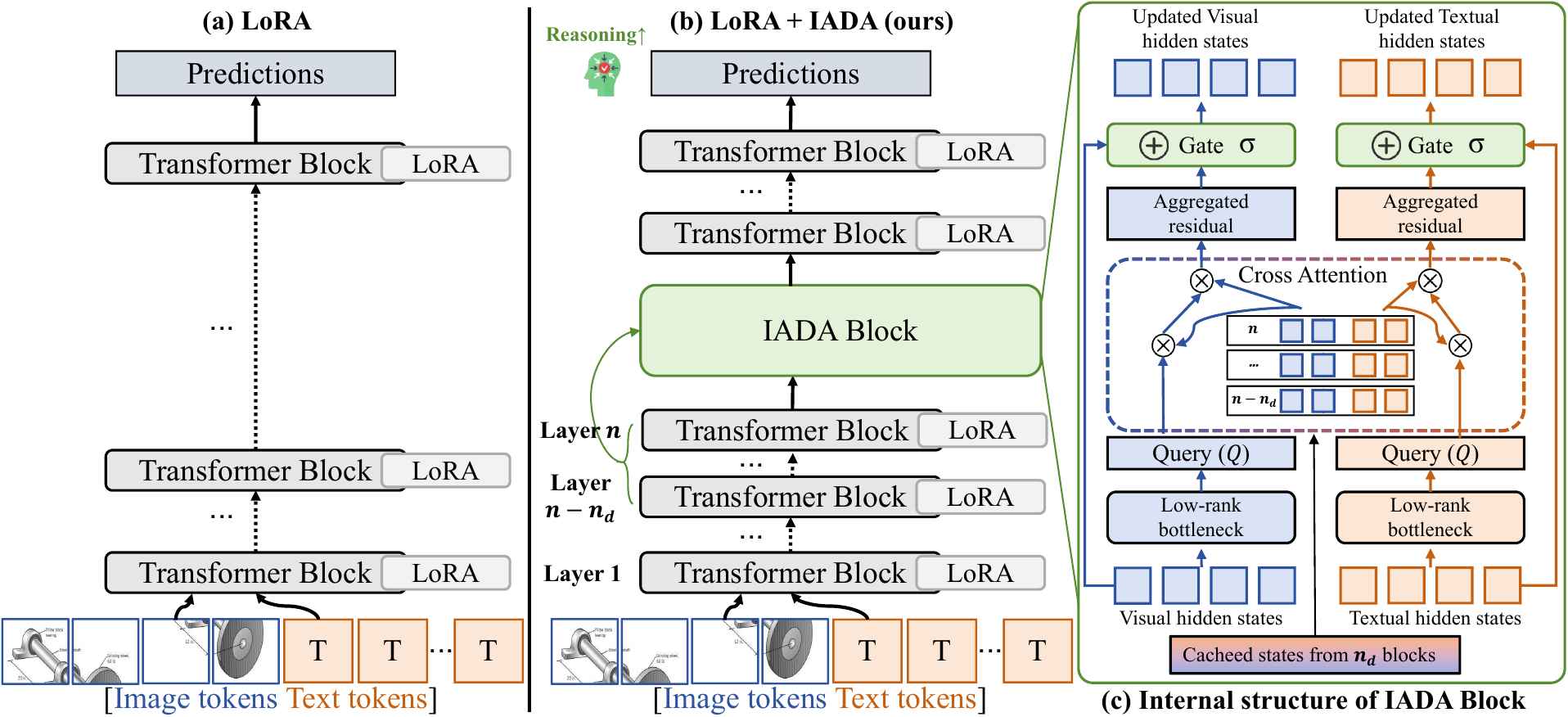}
\caption{
Overview of \method.
Compared with LoRA fine-tuning in (a), \method in (b) inserts an IADA block at block boundaries to enable explicit cross-depth routing.
IADA generates modality-specific queries from the current hidden states, retrieves information from cached states at previous $n_d$ block boundaries, and injects the resulting aggregated residuals back through gated residual addition.
}
\label{fig:overview}
\end{figure}

Motivated by the hypothesis that VLM fine-tuning may weaken access to reasoning-relevant representations across depth, we study cross-depth routing as an independent design axis in multimodal post-training. 
We begin with a fixed-query depth aggregation module as a minimal diagnostic intervention, and then extend it to \method, an input-adaptive and modality-aware routing mechanism that retrieves earlier-layer representations conditioned on the current hidden state. 
As shown in~\cref{fig:overview}, the resulting design is lightweight, structurally complementary to standard PEFT methods, and introduces only a small number of additional parameters.

\subsection{Fixed Depth Aggregation as a Diagnostic Baseline}
\label{ss:fixed_depth}

\nbf{Formulation.}
We first formulate a minimal depth aggregation mechanism to isolate the role of cross-depth representational access. 
Consider a transformer with $L$ layers, partitioned into $K$ blocks of size $B$, such that $L = KB$. 
Let $h^l \in \mathbb{R}^{n \times d}$ denote the hidden state at layer $l$, where $n$ is the sequence length and $d$ is the hidden dimension.

At the end of block $k$ (i.e., layer $l = kB$), we collect the hidden states from all preceding block boundaries,
$\{h^0, h^B, \ldots, h^{(k-1)B}\}$,
and aggregate them through cross-attention:
\begin{equation}
r_k = \mathrm{CrossAttn}\!\left(q,\; [h^0; h^B; \ldots; h^{(k-1)B}] \right),
\label{eq:fixed_attnres}
\end{equation}
where $q \in \mathbb{R}^{1 \times d}$ is a learnable query vector and $[\cdot;\cdot]$ denotes concatenation along the sequence dimension. 
Moreover, $\mathrm{CrossAttn}(\cdot,\cdot)$ denotes standard multi-head cross-attention (details in~\cref{app:cross_attn}).
The aggregated representation is then injected into the current hidden state through a gated residual connection:
\begin{equation}
h^{kB} \leftarrow h^{kB} + \alpha_k \, r_k,
\qquad
\alpha_k = \sigma(\gamma_k),
\label{eq:fixed_gate}
\end{equation}
where $\gamma_k$ is initialized near zero so that the module starts close to an identity mapping.

\nbf{Why it serves as a useful diagnostic.}
This formulation is intended as a diagnostic baseline rather than our final method. 
It restores explicit access to earlier-layer representations without altering the main computations inside each transformer layer, thereby providing a clean testbed for assessing whether cross-depth retrieval helps recover reasoning performance. 
In this sense, it isolates the effect of cross-depth access from other changes to the backbone architecture.

\subsection{Why Fixed Aggregation Is Insufficient}
\label{sec:fixed_limit}

While the fixed-query baseline is useful for diagnosis, it remains limited in two important respects.

\nbf{Input-invariant routing.}
The same learned query is applied to all inputs regardless of the current hidden state. 
This implicitly assumes that all examples should retrieve earlier representations in the same way, even though different inputs may require different forms of cross-depth access.

\nbf{Modality-uniform routing.}
Visual and textual tokens share the same routing mechanism, although the two modalities may exhibit different depth-wise processing patterns. 
For VLMs, this shared routing strategy is restrictive because the representations that should be retrieved for perception and reasoning may not follow the same structure across depth.

These limitations motivate a more expressive formulation in which cross-depth retrieval is treated as a conditional routing problem. 
Instead of using a global query shared across all examples, \method generates aggregation queries from the current hidden state and allows the routing behavior to differ across modalities.

\subsection{Input-Adaptive Depth Aggregation}
\label{sec:iada}

\nbf{Input-adaptive query generation.}
Our first design principle is that the aggregation query should depend on the current input context. 
At each block boundary $k$ and for each modality $m \in \{v,t\}$, we first compute a sequence-level context vector by mean-pooling the current hidden states:
\begin{equation}
c_k^m = \mathrm{MeanPool}(h_{kB}^m) \in \mathbb{R}^{d}.
\label{eq:context_vector}
\end{equation}

We then map this context vector to an aggregation query through a low-rank bottleneck:
\begin{equation}
q_k^m = W_{\mathrm{up}}^m W_{\mathrm{down}}^m c_k^m,
\label{eq:adaptive_query}
\end{equation}
where $W_{\mathrm{down}}^m \in \mathbb{R}^{r \times d}$ and $W_{\mathrm{up}}^m \in \mathbb{R}^{d \times r}$ define a rank-$r$ projection.

We use a sequence-level query rather than token-level queries because the role of \method\ is to control \emph{where} information is retrieved from across depth, rather than to recompute token content. 
A pooled representation therefore provides a compact conditioning signal for global routing decisions while keeping the additional computation small.

\nbf{Modality-specific depth routing.}
Our second design principle is that depth routing should be modality-aware. 
In a VLM, the hidden states at each layer contain both visual and textual tokens. 
We separate them using a binary modality mask and maintain distinct routing streams for the two modalities.

At block boundary $k$, cross-depth aggregation is performed independently as
\begin{equation}
r_k^v = \mathrm{CrossAttn}\!\left(q_k^v,\; [h_0^v; \ldots; h_{(k-1)B}^v] \right),
\label{eq:vision_routing}
\end{equation}
and
\begin{equation}
r_k^t = \mathrm{CrossAttn}\!\left(q_k^t,\; [h_0^t; \ldots; h_{(k-1)B}^t] \right).
\label{eq:text_routing}
\end{equation}
The retrieved representations are then scattered back to their original positions in the full sequence.

Importantly, this design separates routing decisions rather than fully decoupling multimodal processing. 
The backbone still applies shared self-attention over the full interleaved sequence, thereby preserving token-level cross-modal interaction. 
\method only modifies how the model accesses earlier-layer representations, and does so in a modality-aware manner.

\nbf{Low-rank gated injection.}
The bottleneck in~\cref{eq:adaptive_query} is motivated by both efficiency and inductive bias. 
Cross-depth routing is a relatively low-dimensional decision problem: the model does not need to synthesize new token representations, but only to determine which earlier features are useful to retrieve. 
A low-rank parameterization is therefore sufficient to model this routing behavior while keeping the routing module lightweight.

As in the fixed-query baseline, the retrieved representation is injected through a gated residual connection:
\begin{equation}
h_{kB}^m \leftarrow h_{kB}^m + \alpha_k \, r_k^m,
\qquad
\alpha_k = \sigma(\gamma_k),
\label{eq:adaptive_gate}
\end{equation}
where $\gamma_k$ is initialized near zero. 
This keeps the routing pathway close to an identity perturbation at the start of training and improves optimization stability.

\subsection{Relation to PEFT}
\label{sec:relation_peft}

\method is designed to complement rather than replace standard PEFT methods such as LoRA. 
LoRA adapts within-layer transformations by modifying linear operators inside individual transformer blocks. 
In contrast, \method adapts across-depth access by explicitly modeling how the current hidden state retrieves representations from earlier blocks. 
The two mechanisms therefore operate on different adaptation axes and can be combined naturally.

\section{Experiments}
\label{sec:experiments}

\subsection{Experimental Setup}
\label{sec:exp_setup}

We use Qwen3-VL-2B-Instruct~\cite{qwen3vl2025} as our primary backbone. 
The model has $28$ transformer layers with hidden dimension $d=2048$. We partition the network into $K=4$ blocks with block size $B=7$.

Unless otherwise specified, all experiments use LoRA~\cite{hu2022lora} with rank 16 and scaling factor $\alpha=32$, applied to all linear layers. 
Training is performed for one epoch on LLaVA-Instruct-Mix~\cite{liu2024improved}, comprising 259K samples and 16,198 optimization steps with effective batch size $16$.
We optimize LoRA parameters with AdamW~\cite{loshchilov2019decoupled} using learning rate $2 \times 10^{-5}$, and optimize \method parameters with learning rate $1 \times 10^{-3}$. 
Training uses bf16 mixed precision with gradient checkpointing on a single NVIDIA H200 GPU.

We evaluate on 11 benchmarks, grouped into reasoning benchmarks, including MMMU~\cite{yue2024mmmu}, MathVista~\cite{lu2024mathvista}, ScienceQA~\cite{lu2022scienceqa}, MMStar~\cite{chen2024mmstar}, and AI2D~\cite{kembhavi2016diagram}, and perception benchmarks, including MME~\cite{fu2024mme}, POPE~\cite{li2023pope}, RealWorldQA~\cite{li2024realworldqa}, ChartQA~\cite{masry2022chartqa}, TextVQA~\cite{singh2019textvqa}, and OCRBench~\cite{liu2024ocrbench}. 
We report reasoning average (R-Avg), perception average (P-Avg), and overall average (All-Avg) across all 11 benchmarks. MME scores are normalized to the $0\sim100$ range.

We compare four experimental conditions throughout the main experiments: (A) the pre-trained baseline without fine-tuning, (B) standard LoRA supervised fine-tuning, (C) fixed-query attention residuals combined with LoRA, and (D) \method combined with LoRA.





\begin{table}[t]
\centering
\small
\setlength{\tabcolsep}{4pt}
\caption{
Main results on Qwen3-VL-2B across 11 benchmarks. Scores are reported in percentage points; MME is normalized to the 0--100 range. R-Avg and P-Avg denote averages over reasoning and perception benchmarks, respectively. LoRA-only exposes a substantial reasoning tax, fixed depth aggregation recovers much of this loss, and \method achieves the best overall reasoning--perception trade-off among all compared variants in our setting. Best results in each column are shown in bold; ties are bolded.
}
\label{tab:main_results}
\begin{adjustbox}{max width=\textwidth}
\begin{tabular}{l c ccccc cccccc ccc}
\toprule
& \multicolumn{1}{c}{Trainable} & \multicolumn{5}{c}{Reasoning} & \multicolumn{6}{c}{Perception} & \multicolumn{3}{c}{Average} \\
\cmidrule(lr){2-2}
\cmidrule(lr){3-7}
\cmidrule(lr){8-13}
\cmidrule(lr){14-16}
Method & Params & MMMU & MathVista & SciQA & MMStar & AI2D & MME & POPE & RWQ & ChartQA & TextVQA & OCRBench & R-Avg & P-Avg & All-Avg \\
\midrule
A: Pre-trained & 0 & 34.2 & 48.0 & 82.9 & 45.4 & \textbf{76.0} & 82.3 & \textbf{89.3} & \textbf{66.4} & \textbf{79.6} & \textbf{80.2} & 43.1 & 57.3 & 73.5 & 66.1 \\
B: LoRA-only & 7.6M & 31.3 & 42.6 & 72.1 & 44.8 & 70.2 & \textbf{83.2} & 88.6 & 63.4 & 78.0 & 76.2 & 42.5 & 52.2 & 72.0 & 63.0 \\
C: AttnRes+LoRA & 7.7M & 42.2 & 46.1 & 82.6 & 48.4 & 74.0 & 82.7 & 88.0 & 62.6 & 77.9 & 76.1 & 41.1 & 58.7 & 71.4 & 65.6 \\
\rowcolor{iadablue}
D: \method+LoRA & 7.7M & \textbf{44.4} & \textbf{54.0} & \textbf{83.4} & \textbf{51.2} & 75.3 & \textbf{83.2} & 87.5 & 63.9 & 77.9 & 77.6 & \textbf{61.6} & \textbf{61.7} & \textbf{75.3} & \textbf{69.1} \\
\midrule
\multicolumn{2}{l}{\textit{$\Delta$(D$-$A)}} 
& \textit{+10.2} & \textit{+6.0} & \textit{+0.5} & \textit{+5.8} & \textit{--0.7}
& \textit{+0.9} & \textit{--1.8} & \textit{--2.5} & \textit{--1.7} & \textit{--2.6} & \textit{+18.5}
& \textit{+4.4} & \textit{+1.8} & \textit{+3.0} \\
\multicolumn{2}{l}{\textit{$\Delta$(D$-$B)}} 
& \textit{+13.1} & \textit{+11.4} & \textit{+11.3} & \textit{+6.4} & \textit{+5.1}
& \textit{+0.0} & \textit{--1.1} & \textit{+0.5} & \textit{--0.1} & \textit{+1.4} & \textit{+19.1}
& \textit{+9.5} & \textit{+3.3} & \textit{+6.1} \\
\bottomrule
\end{tabular}
\end{adjustbox}
\end{table}

\subsection{Main Results: The Reasoning Tax and Its Recovery}
\label{sec:main_results}

\cref{tab:main_results} summarizes the main results.

\nbf{LoRA-only fine-tuning exposes a substantial reasoning tax.}
Relative to the pre-trained baseline, standard LoRA reduces R-Avg from $57.3$ to $52.2$, with especially large drops on MathVista and ScienceQA. 
By contrast, the gains on perception benchmarks are limited, and the overall average also declines. 
This result indicates that conventional parameter-efficient fine-tuning does not naturally preserve reasoning ability in VLMs, even when it improves task adaptation.

\nbf{Restoring cross-depth access already recovers much of the lost reasoning.}
Adding fixed-query depth aggregation raises R-Avg from $52.2$ to $58.7$, recovering most of the reasoning degradation introduced by LoRA fine-tuning. 
This improvement is substantial given that the underlying transformer computations remain unchanged. 
The result therefore provides direct empirical support for the view that disrupted access to earlier representations is an important factor in the VLM reasoning tax.

\nbf{\method further improves both reasoning and overall perception.}
Building on fixed aggregation, \method reaches $61.7$ R-Avg and $75.3$ P-Avg with only $0.14$M additional parameters, yielding the best overall reasoning--perception trade-off among all compared variants in our setting. 
The gains are especially pronounced on MMMU, MathVista, MMStar, and OCRBench, suggesting that input-adaptive depth routing is particularly beneficial when performance depends on preserving multi-stage abstraction or fine-grained visual-text alignment.

\subsection{Ablations and Design Analysis}

\begin{table}[h]
\centering
\small
\setlength{\tabcolsep}{5pt}
\caption{
From fixed aggregation to adaptive aggregation. The table isolates the main progression from LoRA-only fine-tuning to fixed-query aggregation, adaptive aggregation without modality split, and the full \method design. Fixed aggregation recovers most of the reasoning loss, input adaptivity provides the dominant additional gain, and modality-specific routing yields a smaller final refinement. OCRBench is shown as a representative case where adaptive routing is particularly beneficial.
}
\label{tab:fixed_to_adaptive}
\begin{adjustbox}{max width=\linewidth}
\begin{tabular}{l c c c c c c c c}
\toprule
Method & Query & Routing & AttnRes Params & OCRBench & R-Avg & P-Avg & All-Avg & $\Delta$R-Avg \\
\midrule
B: LoRA-only & -- & -- & 0 & 42.5 & 52.2 & 72.0 & 63.0 & -- \\
C: Fixed AttnRes+LoRA & Fixed & No split & 0.10M & 41.1 & 58.7 & 71.4 & 65.6 & +6.5 \\
Adaptive, No split & Adaptive & No split & 0.07M & 60.5 & 61.2 & 75.6 & 69.0 & +2.5 \\
\rowcolor{iadablue}
D: \method+LoRA & Adaptive & Self-modal & 0.14M & \textbf{61.6} & \textbf{61.7} & 75.3 & \textbf{69.1} & +0.5 \\
\bottomrule
\end{tabular}
\end{adjustbox}
\end{table}

\subsubsection{From Fixed Aggregation to Adaptive Aggregation}
\label{sec:fixed_to_adaptive}

\cref{tab:fixed_to_adaptive} isolates the main progression from LoRA-only fine-tuning to the full \method design. 
Three observations follow.

\nbf{Restoring cross-depth access is the first-order factor.}
Starting from LoRA-only ($52.2$ R-Avg), fixed-query aggregation immediately raises reasoning to $58.7$, recovering most of the loss introduced by fine-tuning. 
This confirms that cross-depth representational access is already a strong baseline for mitigating the reasoning tax.

\nbf{Input adaptivity provides the dominant gain beyond fixed aggregation.}
Replacing fixed queries with input-adaptive queries while keeping a shared routing stream further improves R-Avg from $58.7$ to $61.2$.
This $+2.5$ point gain is substantially larger than the final gain from modality-specific routing, indicating that the main benefit comes from allowing the routing decision to depend on the current input.

\nbf{Modality-specific routing provides a smaller but consistent refinement.}
Adding self-modal routing on top of adaptive aggregation further improves R-Avg from $61.2$ to $61.7$. 
Although this gain is modest, it is consistent with the view that routing decisions benefit from modality-local conditioning. 
At the same time, the strong no-split baseline suggests that modality-specific routing should be understood as a refinement rather than the sole source of improvement.

A representative example is OCRBench. 
Fixed aggregation remains weak on this benchmark ($41.1$), whereas adaptive routing raises performance sharply to $60.5$, and the full \method reaches $61.6$. 
This pattern suggests that conditional cross-depth retrieval is particularly helpful when performance depends on precise visual-text alignment. 
A more fine-grained view of representative intermediate variants is provided in~\cref{app:design_variants}.

\subsubsection{Interaction with LoRA Rank}
\label{sec:lora_interaction}

\cref{tab:lora_interaction} reveals a clear interaction between \method and LoRA capacity.

\nbf{The benefit of \method is much larger at low LoRA rank.}
With LoRA rank $16$, \method improves reasoning by $+9.5$ points and perception by $+3.3$ points. 
When the LoRA rank is increased to $32$, however, the reasoning gain shrinks sharply to only $+0.5$, while the perception gain is reduced to $+2.3$.

\nbf{This pattern suggests that \method contributes more than additional parameter count.}
If \method merely acted as a small increase in capacity, its marginal benefit would not be expected to collapse so substantially when LoRA rank increases. 
Instead, the result suggests partial functional overlap: 
when LoRA capacity is limited, explicit depth routing provides a complementary mechanism that LoRA alone does not easily realize; 
when LoRA rank is larger, some of this routing behavior may already be absorbed implicitly by within-layer adaptation.

\nbf{The practical implication is that \method is most valuable in parameter-efficient regimes.}
The strongest gains arise precisely when LoRA capacity is constrained, which is also the regime in which PEFT methods are most attractive in practice. 
This result therefore positions \method not as a generic replacement for higher-rank adaptation, but as a particularly effective complement when the adaptation budget is small.

\begin{table}[t]
\centering
\small
\caption{
Interaction between \method and LoRA rank. 
LoRA rank and its corresponding trainable parameter count are shown explicitly, where LoRA Params denotes the trainable parameters introduced by LoRA alone. 
The gains of \method are substantially larger at rank $16$ than at rank $32$, indicating that explicit depth routing is especially beneficial when PEFT capacity is constrained.
}
\label{tab:lora_interaction}
\begin{tabular}{c c l c c c}
\toprule
LoRA Rank & LoRA Params & Method & R-Avg & P-Avg & All-Avg \\
\midrule
\multirow{3}{*}{16} & \multirow{3}{*}{3.8M}
& B: LoRA-only & 52.2 & 72.0 & 63.0 \\
& & D: \method+LoRA & \textbf{61.7} & \textbf{75.3} & \textbf{69.1} \\
& & \textit{$\Delta$(D$-$B)} & \textit{+9.5} & \textit{+3.3} & \textit{+6.1} \\
\midrule
\multirow{3}{*}{32} & \multirow{3}{*}{7.6M}
& B: LoRA-only & 54.8 & 72.8 & 64.6 \\
& & D: \method+LoRA & 55.3 & \textbf{75.1} & \textbf{66.1} \\
& & \textit{$\Delta$(D$-$B)} & \textit{+0.5} & \textit{+2.3} & \textit{+1.5} \\
\bottomrule
\end{tabular}
\end{table}

\subsubsection{Query Design and Modality Conditioning}
\label{sec:ablation_query}

\begin{table}[h]
\centering
\small
\setlength{\tabcolsep}{4pt}
\caption{
Ablation on query design and modality conditioning. Group I compares fixed and adaptive queries under different conditioning strategies. Group II isolates modality-specific contributions. Input adaptivity provides the dominant gain, while modality-specific conditioning offers additional but smaller improvements. Full benchmark results are reported in Appendix.
}
\label{tab:ablation_query}
\begin{adjustbox}{max width=\linewidth}
\begin{tabular}{l c c c c c c c c c}
\toprule
Configuration & AttnRes Params & Query & Conditioning & MMMU & MathVista & OCRBench & R-Avg & P-Avg & All-Avg \\
\midrule
\multicolumn{10}{l}{\textit{I. Query type and modality conditioning}} \\
\midrule
\rowcolor{iadablue}
Adaptive, self-modal (ours) & 0.14M & Adaptive & Self-modal & \textbf{44.4} & 54.0 & \textbf{61.6} & \textbf{61.7} & 75.3 & \textbf{69.1} \\
Adaptive, cross-modal & 0.14M & Adaptive & Cross-modal & 39.9 & 51.0 & 56.4 & 59.0 & 74.0 & 67.2 \\
Adaptive, no modality split & 0.07M & Adaptive & Shared & 43.6 & \textbf{54.8} & 60.5 & 61.2 & \textbf{75.6} & 69.0 \\
Fixed query (AttnRes) & 0.10M & Fixed & Shared & 42.2 & 46.1 & 41.1 & 58.7 & 71.4 & 65.6 \\
None (LoRA-only) & 0 & -- & -- & 31.3 & 42.6 & 42.5 & 52.2 & 72.0 & 63.0 \\
\midrule
\multicolumn{10}{l}{\textit{II. Modality-specific contributions}} \\
\midrule
\rowcolor{iadablue}
Both adaptive (ours) & 0.14M & Adaptive & Separate & \textbf{44.4} & \textbf{54.0} & \textbf{61.6} & \textbf{61.7} & \textbf{75.3} & \textbf{69.1} \\
Text-only adaptive & 0.07M & Partial & Text only & 40.8 & 50.3 & 54.8 & 59.6 & 73.5 & 67.2 \\
Visual-only adaptive & 0.07M & Partial & Vision only & 40.1 & 46.3 & 46.4 & 58.1 & 72.5 & 66.0 \\
Shared projection & 0.07M & Adaptive & Shared projection & 42.0 & 52.4 & 60.7 & 59.7 & 74.7 & 67.9 \\
\bottomrule
\end{tabular}
\end{adjustbox}
\end{table}

\cref{tab:ablation_query} examines two design axes jointly: query generation and modality conditioning.

\nbf{Input adaptivity is the dominant factor.}
Compared with fixed queries, adaptive self-modal queries improve R-Avg from $58.7$ to $61.7$, a gain of $+3.0$ points. 
The improvement is concentrated on reasoning-heavy benchmarks such as MathVista and MMMU, and is especially pronounced on OCRBench. 
This pattern indicates that different inputs require different depth-routing strategies, and that a fixed query is insufficient to capture this variability.

\nbf{Modality-specific conditioning provides additional but smaller gains.}
Self-modal conditioning improves over cross-modal conditioning by $+2.7$ points in R-Avg, suggesting that modality-local signals are more effective for routing decisions in this setting. 
At the same time, the no-split variant remains highly competitive at $61.2$ R-Avg and $69.0$ All-Avg. 
This result suggests that the primary benefit comes from input adaptivity, while modality-specific routing is better understood as a refinement rather than the sole source of improvement.

\nbf{Text-side adaptivity appears more important than vision-side adaptivity.}
When only one modality uses adaptive queries, text-only adaptation outperforms vision-only adaptation ($59.6$ vs.\ $58.1$ R-Avg). 
This pattern suggests that linguistic reasoning is more sensitive to routing flexibility than visual pattern extraction. 
In addition, using separate projections outperforms a shared projection, indicating that modality-specific parameterization still provides a useful inductive bias even when the overall routing mechanism is shared.

\subsubsection{Low-Rank Bottleneck}
\label{sec:ablation_rank}

\begin{table}[t]
\centering
\small
\caption{
Effect of bottleneck rank on performance and parameter count. A rank-16 bottleneck achieves the best reasoning performance while using the fewest additional parameters, suggesting that effective depth routing can be captured with a surprisingly compact parameterization in our setting. Full benchmark results are reported in Appendix.
}
\label{tab:ablation_rank}
\begin{adjustbox}{max width=\linewidth}
\begin{tabular}{l c c c c c c c c}
\toprule
Rank & AttnRes Params & Total Params & MMMU & MathVista & OCRBench & R-Avg & P-Avg & All-Avg \\
\midrule
\rowcolor{iadablue}
$r=16$ (ours) & 0.14M & 7.7M & \textbf{44.4} & 54.0 & 61.6 & \textbf{61.7} & 75.3 & 69.1 \\
$r=64$ & 0.53M & 8.1M & 42.7 & 52.8 & \textbf{63.0} & 60.2 & 75.0 & 68.3 \\
$r=256$ & 2.1M & 9.7M & 41.7 & 55.6 & 62.0 & 61.5 & \textbf{75.8} & \textbf{69.3} \\
Full-rank ($r=d$) & 8.4M & 16.0M & 43.2 & \textbf{55.7} & 60.0 & 60.4 & 75.3 & 68.5 \\
\bottomrule
\end{tabular}
\end{adjustbox}
\end{table}

\cref{tab:ablation_rank} studies the effect of bottleneck rank on routing performance and parameter count.

\nbf{A very small bottleneck is already sufficient for strong reasoning performance.}
The rank-16 bottleneck achieves the best reasoning score ($61.7$ R-Avg) while introducing only $0.14$M additional parameters. 
It matches or exceeds substantially larger configurations, including the full-rank variant, which uses $8.4$M routing parameters but reaches only $60.4$ R-Avg. 
This corresponds to roughly $60\times$ fewer routing parameters with a stronger reasoning result.

\nbf{Increasing the rank does not lead to monotonic improvement.}
Performance does not consistently improve as the bottleneck becomes larger: $r=16$ achieves the highest R-Avg, followed by $r=256$, full-rank, and $r=64$. 
This non-monotonic pattern suggests that a larger routing space is not necessarily easier to optimize, and that the bottleneck may act as a useful regularizer rather than merely a compression device.

\nbf{The low-rank result points to an efficient routing parameterization.}
We therefore interpret the bottleneck ablation as evidence that useful depth-routing behavior can be captured with a surprisingly compact parameterization in our setting. 
This observation is consistent with the broader idea that many adaptation problems admit low-dimensional structure~\cite{aghajanyan2021intrinsic, li2018measuring}. 
In our setting, the bottleneck constrains cross-depth routing rather than transformation content.

\subsubsection{Extended Design Exploration}
\label{sec:extended}

\begin{table}[h]
\centering
\small
\setlength{\tabcolsep}{4pt}
\caption{
Extended design exploration beyond the default \method configuration. Sequence-level routing, a linear bottleneck, and a globally learned gate remain preferable in our setting. Even under extreme compression ($r=4$), \method still improves over LoRA-only.
}
\label{tab:extended_main}
\begin{adjustbox}{max width=\linewidth}
\begin{tabular}{l c c c c c}
\toprule
Configuration & AttnRes Params & R-Avg & P-Avg & All-Avg & $\Delta$All vs.\ LoRA \\
\midrule
LoRA-only & 0 & 52.2 & 72.0 & 63.0 & -- \\
\rowcolor{iadablue}
\method (default) & 0.14M & \textbf{61.7} & \textbf{75.3} & \textbf{69.1} & \textbf{+6.1} \\
\midrule
Token-level routing, $r=16$ & 0.14M & 59.4 & 74.9 & 67.9 & +4.9 \\
Nonlinear bottleneck, $r=16$ & 0.14M & 58.5 & 72.3 & 66.0 & +3.0 \\
Adaptive gate, $r=16$ & 0.14M & 55.6 & 70.9 & 64.0 & +1.0 \\
Extra-small rank, $r=4$ & 0.04M & 59.0 & 73.2 & 66.7 & +3.7 \\
\bottomrule
\end{tabular}
\end{adjustbox}
\end{table}

Table~\ref{tab:extended_main} summarizes several secondary design variations beyond the default \method configuration.

\nbf{Sequence-level routing is preferable to token-level routing.}
Replacing sequence-level queries with token-level routing reduces R-Avg from $61.7$ to $59.4$. 
This result suggests that depth aggregation is better treated as a global routing decision for the current input, rather than as token-wise content recomputation. 
A more fine-grained comparison, including token-level full-rank routing, is reported in~\cref{app:appendix_extended}.

\nbf{A linear bottleneck provides a stronger inductive bias than a nonlinear one.}
Inserting a GeLU nonlinearity into the bottleneck lowers R-Avg to $58.5$ and All-Avg to $66.0$. 
This suggests that, in our setting, depth routing is better modeled by a compact linear parameterization than by a more expressive nonlinear mapping.

\nbf{A globally learned gate is more effective than an input-dependent gate.}
Making the residual gate input-adaptive reduces R-Avg to $55.6$ and All-Avg to $64.0$. 
This result suggests that per-input gate modulation introduces unnecessary instability, whereas a globally learned gate provides a more reliable way to control the strength of cross-depth aggregation.

\nbf{Extreme compression still preserves a substantial portion of the gain.}
Even with rank $r=4$, \method remains clearly above LoRA-only, improving All-Avg by $+3.7$ points. 
Meanwhile, the default rank-16 configuration remains the strongest overall setting, indicating that the routing module admits aggressive compression but still benefits from a small amount of additional capacity.

\section{Discussion}
\label{sec:discussion}

\nbf{Across-layer routing as a distinct adaptation axis.}
A central implication of our results is that multimodal fine-tuning should not be viewed solely through the lens of within-layer adaptation. 
Existing PEFT methods, including LoRA~\cite{hu2022lora}, primarily modify transformations inside individual layers. 
By contrast, \method operates on a different axis: it changes how the model accesses and reuses representations across depth. 
In this sense, our contribution is not merely another PEFT variant, but evidence that \emph{across-layer routing} is itself a meaningful design dimension in VLM fine-tuning. 
The fact that even fixed depth aggregation recovers a large fraction of the reasoning loss suggests that preserving access to earlier representations is already a strong intervention, independent of any additional adaptivity.

\nbf{What the current evidence most strongly supports.}
Our experiments support three conclusions most directly. 
First, preserving cross-depth representational access matters: fixed aggregation already recovers much of the reasoning degraded by LoRA fine-tuning. 
Second, making this access input-adaptive matters further: the largest additional gain beyond fixed aggregation comes from allowing routing decisions to depend on the current input. 
Third, this mechanism is especially useful in parameter-efficient regimes: the gain of \method is much larger when LoRA capacity is limited than when LoRA rank is increased. 
At the same time, several findings should be interpreted with appropriate caution. 
The strong no-modality-split baseline suggests that modality-specific routing is better understood as a useful refinement than as the sole source of improvement. 
Similarly, the bottleneck-rank results suggest that effective routing admits a highly compact parameterization in our setting, but they do not by themselves establish a general intrinsic-dimensionality law for depth routing.

\nbf{Scope and limitations.}
The present results should be interpreted within the scope of our experimental setting. 
All experiments are conducted on a single backbone (Qwen3-VL-2B) and a single instruction-tuning dataset, so generalization across model scales, architecture families, and data distributions remains to be established. 
Our routing design also relies on a fixed modality mask, which may become suboptimal in deeper layers where visual and textual representations are more entangled. 
In addition, our evaluation focuses on standard benchmark performance; the effect of depth routing on open-ended generation quality and free-form multimodal interaction is not measured here. 
These limitations clarify the intended contribution of this work: not a definitive mechanism for all VLM fine-tuning settings, but a strong empirical case that preserving and adapting cross-depth access is an important and previously underexplored direction.

\section{Conclusion}
\label{sec:conclusion}

We investigated a plausible explanation for the reasoning tax in VLM fine-tuning: that supervised adaptation may disrupt access to depth-wise representations that support reasoning. 
Our results suggest that preserving cross-depth representational access is an important missing factor during fine-tuning, and that making this access input-adaptive yields further gains.
Based on this observation, we proposed \method, a lightweight mechanism for input-adaptive depth aggregation. 
With only $0.14$M additional parameters, \method substantially improves reasoning while also improving overall perception, with the strongest gains appearing under low-rank PEFT. 
Our ablations further indicate that input adaptivity is the main driver of the improvement, while modality-specific routing and low-rank parameterization provide useful additional structure.
More broadly, our findings suggest that multimodal fine-tuning should be studied not only through within-layer adaptation, but also through how information is routed across depth. 
We hope this work motivates further investigation into across-layer adaptation as an independent design axis in VLM post-training.


\bibliographystyle{unsrtnat}
\bibliography{references}

\clearpage
\appendix
\section*{Appendix}

\section{Details of \method}

\subsection{Definition of Cross-Depth Cross-Attention}
\label{app:cross_attn}

For completeness, we define the cross-attention operator used in Equations.~\eqref{eq:fixed_attnres}, \eqref{eq:vision_routing}, and \eqref{eq:text_routing}. 
Given a query matrix $Q \in \mathbb{R}^{n_q \times d}$ and a source sequence
$X \in \mathbb{R}^{n_x \times d}$, cross-attention is defined as
\begin{equation}
\mathrm{CrossAttn}(Q, X)
=
\mathrm{Concat}(\mathrm{head}_1, \ldots, \mathrm{head}_H) W_O,
\end{equation}
where each head is computed as
\begin{equation}
\mathrm{head}_i
=
\mathrm{softmax}
\left(
\frac{(QW_Q^{(i)})(XW_K^{(i)})^\top}{\sqrt{d_h}}
\right)
(XW_V^{(i)}),
\end{equation}
with $H$ attention heads, head dimension $d_h = d/H$, and learned projection
matrices $W_Q^{(i)}, W_K^{(i)}, W_V^{(i)} \in \mathbb{R}^{d \times d_h}$ and
$W_O \in \mathbb{R}^{d \times d}$.

In our setting, the query is a sequence-level routing vector generated from the
current hidden state at block boundary $k$. For modality $m \in \{v,t\}$, we use
\begin{equation}
q_k^m \in \mathbb{R}^{1 \times d},
\end{equation}
and define the cross-depth memory as the concatenation of hidden states from all
preceding block boundaries:
\begin{equation}
X_k^m = [h_0^m; h_B^m; \ldots; h_{(k-1)B}^m].
\end{equation}
The resulting aggregation is
\begin{equation}
r_k^m = \mathrm{CrossAttn}(q_k^m, X_k^m) \in \mathbb{R}^{1 \times d}.
\end{equation}
Thus, the operator performs retrieval over cached representations across depth,
rather than token mixing within the current layer.

For the fixed-query baseline in~\cref{eq:fixed_attnres}, the same definition
applies, except that the query is replaced by a global learnable vector
$q \in \mathbb{R}^{1 \times d}$ shared across inputs.

\subsection{Modality Mask and Scatter-Back}
\label{app:scatter_back}

At each layer, the hidden state contains both visual and textual tokens. 
We use a binary modality mask to separate them into two modality-specific subsequences,
denoted by $h_{kB}^v$ and $h_{kB}^t$. Cross-depth aggregation is then performed
independently for the two modalities, yielding $r_k^v$ and $r_k^t$.

Since each routing output is a sequence-level vector, we inject it back to the
corresponding modality stream by broadcasting it over the token positions selected
by the modality mask. Concretely, for modality $m \in \{v,t\}$, let
$\mathcal{I}^m$ denote the token indices belonging to that modality. We update
the hidden states at block boundary $kB$ as
\begin{equation}
h_{kB,j} \leftarrow h_{kB,j} + \alpha_k r_k^m,
\qquad j \in \mathcal{I}^m,
\end{equation}
where $\alpha_k = \sigma(\gamma_k)$ is the learnable gate defined in~\cref{eq:adaptive_gate}. In other words, the aggregated routing signal is
scattered back to the original positions of the corresponding modality in the
full interleaved sequence.

This design preserves the original token ordering and keeps token-level
cross-modal interaction unchanged in the backbone, while allowing the routing
signal itself to be conditioned on modality.


\section{Additional Analysis}
\label{app:add_analaysis}

\subsection{Additional Design Variants}
\label{app:design_variants}

\begin{table}[t]
\centering
\small
\setlength{\tabcolsep}{4pt}
\caption{
Representative design variants along the path from fixed aggregation to the final \method design. The table includes intermediate configurations explored during development, highlighting the effects of modality splitting, per-block gating, and input-adaptive query generation.
}
\label{tab:design_path}
\begin{adjustbox}{max width=\textwidth}
\begin{tabular}{l ccccc cccccc ccc}
\toprule
& \multicolumn{5}{c}{Reasoning} & \multicolumn{6}{c}{Perception} & \multicolumn{3}{c}{Average} \\
\cmidrule(lr){2-6}
\cmidrule(lr){7-12}
\cmidrule(lr){13-15}
Configuration & MMMU & MathVista & SciQA & MMStar & AI2D & MME & POPE & RWQ & ChartQA & TextVQA & OCRBench & R-Avg & P-Avg & All-Avg \\
\midrule
LoRA-only & 31.3 & 42.6 & 72.1 & 44.8 & 70.2 & 83.2 & 88.6 & 63.4 & 78.0 & 76.2 & 42.5 & 52.2 & 72.0 & 63.0 \\
Fixed query, modality-aware & 43.7 & 49.5 & 82.4 & 49.3 & 74.4 & 79.3 & 89.3 & 62.9 & 78.2 & 76.6 & 33.7 & 59.9 & 70.0 & 65.4 \\
+ Per-block gate & 43.7 & 48.1 & 82.8 & 49.6 & 74.4 & 81.8 & 88.9 & 62.1 & 78.0 & 76.5 & 38.1 & 59.7 & 70.9 & 65.8 \\
Fixed query, no modality split & 42.2 & 46.1 & 82.6 & 48.4 & 74.0 & 82.7 & 88.0 & 62.6 & 77.9 & 76.1 & 41.1 & 58.7 & 71.4 & 65.6 \\
\rowcolor{iadablue}
Adaptive query, self-modal, rank 16 (\method) & 44.4 & 54.0 & 83.4 & 51.2 & 75.3 & 83.2 & 87.5 & 63.9 & 77.9 & 77.6 & 61.6 & 61.7 & 75.3 & 69.1 \\
\bottomrule
\end{tabular}
\end{adjustbox}
\end{table}

To complement the main ablations, we report several representative intermediate variants explored during the development of \method in~\cref{tab:design_path}. 
These variants are intended to provide additional context on the design path from fixed aggregation to the final model. 
They should not be interpreted as a strictly linear sequence in which each row differs from the previous one by exactly a single modification. 
Instead, they summarize several representative configurations that help clarify the role of modality splitting, per-block gating, and input-adaptive query generation.

Several observations are worth noting. 
First, fixed-query aggregation already recovers a substantial portion of the reasoning loss relative to LoRA-only, consistent with the main text. 
Second, adding a per-block gate alone does not materially change the overall picture, suggesting that gating improves stability but is not the main driver of the gain. 
Third, the final jump to \method is primarily associated with input-adaptive query generation, which substantially improves both reasoning and OCRBench performance. 
Overall, these intermediate variants support the interpretation that the main contribution of \method lies in making cross-depth routing conditional on the current input, while other design choices play a secondary but still useful role.


\subsection{Additional Extended Variants}
\label{app:appendix_extended}

\begin{table*}[h]
\centering
\small
\setlength{\tabcolsep}{5pt}
\caption{
Supplementary results for additional architectural variants discussed in~\cref{sec:extended}. All configurations use LoRA rank 16 as the backbone unless otherwise noted. We report only variants with complete evaluation results.
}
\label{tab:extended_full}
\begin{adjustbox}{max width=\textwidth}
\begin{tabular}{l c c c c c c}
\toprule
Configuration & AttnRes Params & Total Params & R-Avg & P-Avg & All-Avg & $\Delta$All vs.\ LoRA \\
\midrule
\rowcolor{iadablue}
\method\ (self-modal, $r=16$) & 0.14M & 7.7M & \textbf{61.7} & 75.3 & \textbf{69.1} & \textbf{+6.1} \\
\midrule
\multicolumn{7}{l}{\textit{Token-level routing}} \\
Token-level routing, $r=16$ & 0.14M & 7.7M & 59.4 & 74.9 & 67.9 & +4.9 \\
Token-level routing, full-rank & 8.4M & 16.0M & 58.5 & 74.2 & 67.1 & +4.1 \\
\midrule
\multicolumn{7}{l}{\textit{Nonlinear bottleneck}} \\
Nonlinear bottleneck, $r=16$ & 0.14M & 7.7M & 58.5 & 72.3 & 66.0 & +3.0 \\
Nonlinear bottleneck, $r=64$ & 0.53M & 8.1M & 57.9 & \textbf{75.5} & 67.5 & +4.5 \\
\midrule
\multicolumn{7}{l}{\textit{Alternative pooling}} \\
Attention pooling, $r=16$ & 0.14M & 7.7M & 59.8 & 75.1 & 68.1 & +5.1 \\
Attention pooling + no modality split, $r=16$ & 0.14M & 7.7M & 61.2 & 75.3 & 68.9 & +5.9 \\
\midrule
\multicolumn{7}{l}{\textit{Adaptive gate}} \\
Adaptive gate, $r=16$ & 0.14M & 7.7M & 55.6 & 70.9 & 64.0 & +1.0 \\
\midrule
\multicolumn{7}{l}{\textit{Extreme compression}} \\
Extra-small rank, $r=4$ & 0.04M & 7.6M & 59.0 & 73.2 & 66.7 & +3.7 \\
Extra-small rank, $r=8$ & 0.07M & 7.7M & 56.1 & 72.0 & 64.8 & +1.8 \\
\bottomrule
\end{tabular}
\end{adjustbox}
\end{table*}

To complement the compact comparison in~\cref{sec:extended},~\cref{tab:extended_full} reports a broader set of fully evaluated architectural variants. 
These experiments are intended to clarify several secondary design choices, including token-level versus sequence-level routing, linear versus nonlinear bottlenecks, alternative pooling strategies, adaptive gating, and extreme low-rank compression. 
For clarity, we report only variants with complete evaluation results.


\subsection{Complete Benchmark Results}
\label{sec:appendix_full}

\cref{tab:full_results} reports the complete benchmark results for all fully evaluated configurations discussed in the main text and appendix. 
Unlike the compact tables in~\cref{sec:experiments}, which are organized around specific design questions, this table is intended as a comprehensive reference. 
To facilitate cross-configuration comparison, all rows are sorted by overall average (All-Avg) in descending order. 
This table complements the main-text ablations by providing the full benchmark breakdown for each representative variant.

\begin{table}[t]
\centering
\small
\setlength{\tabcolsep}{4pt}
\caption{
Complete benchmark results for all fully evaluated configurations on Qwen3-VL-2B. Scores are reported in percentage points; MME is normalized to the 0--100 range. Rows are sorted by overall average (All-Avg) in descending order. Best results in each column are shown in bold; ties are bolded.
}
\label{tab:full_results}
\begin{adjustbox}{max width=\textwidth}
\begin{tabular}{l ccccc cccccc ccc}
\toprule
& \multicolumn{5}{c}{Reasoning} & \multicolumn{6}{c}{Perception} & \multicolumn{3}{c}{Average} \\
\cmidrule(lr){2-6}
\cmidrule(lr){7-12}
\cmidrule(lr){13-15}
Configuration & MMMU & MathVista & SciQA & MMStar & AI2D & MME & POPE & RWQ & ChartQA & TextVQA & OCRBench & R-Avg & P-Avg & All-Avg \\
\midrule
Self-modal, $r=256$ & 41.7 & 55.6 & 82.0 & 52.1 & \textbf{76.2} & 83.6 & 88.7 & 64.7 & 78.6 & 77.1 & \textbf{62.0} & 61.5 & \textbf{75.8} & \textbf{69.3} \\
\rowcolor{iadablue}
\method\ (self-modal, $r=16$) & \textbf{44.4} & 54.0 & \textbf{83.4} & 51.2 & 75.3 & 83.2 & 87.5 & 63.9 & 77.9 & 77.6 & 61.6 & \textbf{61.7} & 75.3 & 69.1 \\
No modality split & 43.6 & 54.8 & 81.1 & 52.3 & 74.1 & \textbf{84.0} & 88.0 & 64.7 & 78.7 & \textbf{77.8} & 60.5 & 61.2 & 75.6 & 69.0 \\
Cross-modal, $r=64$ & 43.0 & \textbf{56.1} & 81.1 & \textbf{52.5} & 75.3 & 83.1 & 88.6 & 63.5 & 78.2 & 77.3 & 55.3 & 61.6 & 74.3 & 68.5 \\
Self-modal, full-rank & 43.2 & 55.7 & 79.9 & 49.9 & 73.5 & 82.9 & 87.3 & \textbf{65.0} & \textbf{78.9} & \textbf{77.8} & 60.0 & 60.4 & 75.3 & 68.5 \\
Self-modal, $r=64$ & 42.7 & 52.8 & 80.8 & 50.5 & 74.2 & 80.5 & 89.2 & 62.1 & 78.0 & 77.5 & \textbf{63.0} & 60.2 & 75.0 & 68.3 \\
Shared projection & 42.0 & 52.4 & 81.8 & 47.8 & 74.3 & 82.5 & 86.6 & 63.5 & 78.1 & 76.9 & 60.7 & 59.7 & 74.7 & 67.9 \\
Cross-modal, $r=16$ & 39.9 & 51.0 & 80.9 & 49.3 & 74.1 & 83.8 & 85.9 & 62.7 & 78.1 & 76.9 & 56.4 & 59.0 & 74.0 & 67.2 \\
Text-only adaptive & 40.8 & 50.3 & 81.5 & 50.7 & 74.7 & 80.9 & 87.1 & 63.4 & 78.1 & 76.7 & 54.8 & 59.6 & 73.5 & 67.2 \\
A: Pre-trained & 34.2 & 48.0 & 82.9 & 45.4 & 76.0 & 82.3 & \textbf{89.3} & \textbf{66.4} & 79.6 & 80.2 & 43.1 & 57.3 & 73.5 & 66.1 \\
Vision-only adaptive & 40.1 & 46.3 & 82.3 & 48.0 & 74.0 & 81.3 & 89.1 & 63.7 & 77.8 & 77.1 & 46.4 & 58.1 & 72.5 & 66.0 \\
Per-block gate & 43.7 & 48.1 & 82.8 & 49.6 & 74.4 & 81.8 & 88.9 & 62.1 & 78.0 & 76.5 & 38.1 & 59.7 & 70.9 & 65.8 \\
C: Fixed AttnRes & 42.2 & 46.1 & 82.6 & 48.4 & 74.0 & 82.7 & 88.0 & 62.6 & 77.9 & 76.1 & 41.1 & 58.7 & 71.4 & 65.6 \\
Modality-aware fixed query & 43.7 & 49.5 & 82.4 & 49.3 & 74.4 & 79.3 & \textbf{89.3} & 62.9 & 78.2 & 76.6 & 33.7 & 59.9 & 70.0 & 65.4 \\
B: LoRA-only & 31.3 & 42.6 & 72.1 & 44.8 & 70.2 & 83.2 & 88.6 & 63.4 & 78.0 & 76.2 & 42.5 & 52.2 & 72.0 & 63.0 \\
\bottomrule
\end{tabular}
\end{adjustbox}
\end{table}

Several broader patterns are visible from~\cref{tab:full_results}.

\nbf{The strongest reasoning performance and the highest overall average do not fully coincide.}
The default \method configuration achieves the highest reasoning average ($61.7$), whereas the self-modal $r=256$ variant slightly improves overall average and perception average. 
This pattern reinforces the interpretation in the main text that the bottleneck rank trades off reasoning strength, perception strength, and parameter efficiency, rather than improving all dimensions monotonically.

\nbf{Input adaptivity is the primary source of improvement, while modality-specific routing is a refinement.}
The no-modality-split variant remains highly competitive, reaching $61.2$ R-Avg and $69.0$ All-Avg. 
This supports the main-text conclusion that input-adaptive routing is the dominant factor, whereas modality-specific routing provides additional but smaller gains.

\nbf{OCRBench is particularly sensitive to adaptive routing.}
Variants with adaptive queries consistently outperform fixed-query baselines on OCRBench by a large margin. 
In contrast, fixed-query variants remain relatively weak on this benchmark, even when they are competitive on several reasoning metrics. 
This pattern further supports the view that conditional cross-depth retrieval is especially beneficial when performance depends on fine-grained visual-text alignment.

\nbf{The reasoning tax remains substantial across the full comparison set.}
LoRA-only remains near the bottom of the ranking, substantially below most adaptive-routing variants.
This confirms that the degradation induced by standard fine-tuning is not a marginal effect, but a strong and persistent failure mode across the broader design space we explored.

\end{document}